%% file: main.tex
\pdfoutput=1
\documentclass[11pt]{article}
\input{preamble}
\hypersetup{pdftitle={\papertitle}}

\title{\papertitle}
\paperauthors
\date{}

\begin{document}
\maketitle

\begin{abstract}
\noindent
Video recorded on farms, in factories and in laboratories has value beyond the study it was made for:
labelled, it trains behaviour and event detectors; unlabelled, it can pre-train video encoders and world
models for its domain. The same footage shows staff and visitors, and it cannot be released until they
are hidden. Face blurring does not hide them: people at work are seen from above,
from behind, or only as a pair of boots behind a pen wall. Whole-body detectors either miss people or
cover the animals under study, and nobody can look through millions of frames. OpenVeil is an open-source
tool for this problem. Grounding DINO, an open-vocabulary detector, proposes person boxes at a threshold
low enough to miss almost no one, and a DINOv3 crop classifier adapted with low-rank adaptation (LoRA)
rejects the false proposals. A local review tool then turns a reviewer's decisions into corrections. We
used OpenVeil to mask a public calf-behaviour dataset of 1,173 videos (50.8 million frames) from two
dairy farms. The detector ran on 5\% of frames, and the logged review took one author about 2.4~h.
On 731 human-checked frames, OpenVeil covered all 172 people, with a false mosaic box on 3.3\% of
frames. When the alternative detectors were limited to the same false-box rate, the best of them covered
126 people, and Grounding DINO alone covered 71. A face detector found a face for only 66 people. The
boxes of this test set were drawn while checking the pipeline's own proposals.
Code and verifier weights are public.
\end{abstract}

\noindent\textbf{Keywords:} privacy; video anonymisation; person detection; open-vocabulary detection;
DINOv3; low-rank adaptation; human-in-the-loop review; open data

\section{Introduction}
\label{sec:intro}

Video is becoming a resource in its own right. Labelled recordings train models that recognise behaviour,
count events and raise alerts; in livestock farming, cameras already monitor feeding, lying, lameness and
social behaviour \citep{oliveira2021livestock,li2024cowbehavior,bergamini2021pig}. Unlabelled recordings
are useful too. Self-supervised video models learn representations from raw clips that transfer to new
tasks with few labels \citep{tong2022videomae,bardes2024vjepa}, and further self-supervised pre-training
on in-domain images improves a general model on that domain \citep{reed2022hpt}. Models that predict
future frames can flag what they fail to predict as an anomaly \citep{liu2018futureframe} or serve as
world models for training and planning control policies
\citep{ha2018worldmodels,hafner2025dreamerv3,assran2025vjepa2}. Each of these uses needs large amounts of
footage from the domain itself.

The parties that record such footage, from research and commercial farms to industrial sites,
increasingly share it \citep{ragusa2021meccano,vu2024mmcows,yang2026calfdataset}. Every release raises
the question of the staff, visitors and bystanders who were filmed. Images of identifiable people are
personal data under the EU General Data Protection Regulation \citepalias[Art.~4(1)]{gdpr2016}, and
somebody who walks past a barn or factory camera has rarely agreed to appear in a public dataset. Whether
more datasets can be released thus depends on removing people from them reliably, at scale and without
destroying what the data were recorded for.

Current practice does not achieve all three. Face blurring, the most common measure
\citep{frome2009streetview,raina2023egoblur,yang2022imagenetfaces} and the main subject of the literature
on visual privacy \citep{padillalopez2015survey,ribaric2016survey}, assumes a visible and detectable
face, which working footage often does not offer. A blurred face does not prevent recognition either:
people are recognised from pixelated faces, bodies, clothing and context
\citep{mcpherson2016defeating,oh2016faceless,dietlmeier2021faces}. Whole-body detection is the obvious
alternative, but a general detector such as OWLv2 \citep{minderer2023owlv2} or Grounding DINO
\citep{liu2023groundingdino} offers a single score threshold. Where animals or objects resemble a
crouching or half-hidden person, that threshold either misses people or masks the subject of the
recording (\cref{sec:coverage}). Replacing whole bodies with generated ones
\citep{hukkelas2023deepprivacy2,maximov2020ciagan} inherits the difficulty, because it also needs a
detector to say where to act. Manual review does not scale: in one partner's de-identification pipeline
for Ego4D, 500~h of video took 780~h of manual work \citep[supplementary Appendix~B]{grauman2022ego4d},
and a dataset of 50 million frames, the size of the case study below, would take more than 8,000~h at an
optimistic 0.6~s per frame.
Large datasets have instead been de-identified by automatic detection followed by human correction
\citep{caesar2020nuscenes,grauman2022ego4d}. Camera-trap researchers report that people are often
captured by accident \citep{sandbrook2018bycatch}; MegaDetector finds such images but does not mask them
\citep{beery2019megadetector,hernandez2024pytorchwildlife}.

The parts of a better answer exist. Proposing regions and then classifying each one is a classical
pattern \citep{viola2001cascade,girshick2014rcnn}; self-supervised backbones such as DINOv2 and DINOv3
adapt to new tasks with few labels \citep{oquab2024dinov2,simeoni2025dinov3}, also through
parameter-efficient fine-tuning \citep{hu2022lora,yang2026peft}; and verifying a proposed box costs a
person far less than drawing one \citep{papadopoulos2016verification}, which annotation tools such as
CVAT \citep{cvat2020} support but do not organise for privacy review. We are not aware of an open tool
and a documented case study for full-body masking of people in long continuous livestock video, where
the animals are what a detector most often takes for people.

OpenVeil \citep{yang2026openveil} is a tool for footage in which people have to be hidden and something
else is the subject. An open-vocabulary detector, Grounding DINO, proposes person boxes at a low score
threshold, so that almost no person is missed. A crop classifier built on DINOv3 and adapted with
low-rank adaptation (LoRA) \citep{hu2022lora} then verifies each proposal and rejects the false ones. A
person reviews the frames where the pipeline is most likely to be wrong, in a tool that turns each
decision into a correction. Detection stores boxes and leaves the pixels untouched, so every later rule,
margin and correction can be applied again without a graphics processing unit (GPU).

We describe OpenVeil through the release it was built for, a public dataset of calf play behaviour
\citep{yang2026calfdataset,yang2026calfplay}. The software, with the review tool, is under the
Apache-2.0 licence, and the verifier weights are public \citep{yang2026verifier}. Each setting rests on a
measurement: we compare the pipeline with single detectors and a face detector, and its verifier,
threshold, margins and scan schedule with alternatives. The release is the worked example: all 1,173
videos and 1.1 million extracted frames of the dataset were masked, checked and published. Section,
table and figure numbers that begin with S refer to the Supplementary Material.

\section{OpenVeil: propose, verify, review}
\label{sec:openveil}

\subsection{Three commands around one box table}
\label{sec:overview}

OpenVeil starts from the view that a missed person costs more than a false mosaic. Detection is built
for recall, and finding people is kept apart from deciding which boxes are people: one model proposes
generously, and a second judges each proposal. What detection writes is a small table of boxes for each
video, and what is hidden in the release is computed from that table afterwards. The box table is thus
the audit record of a release, and every rule that follows detection can be changed and applied again
without running a model.

\begin{figure}[tb]
  \centering
  \includegraphics[width=\linewidth]{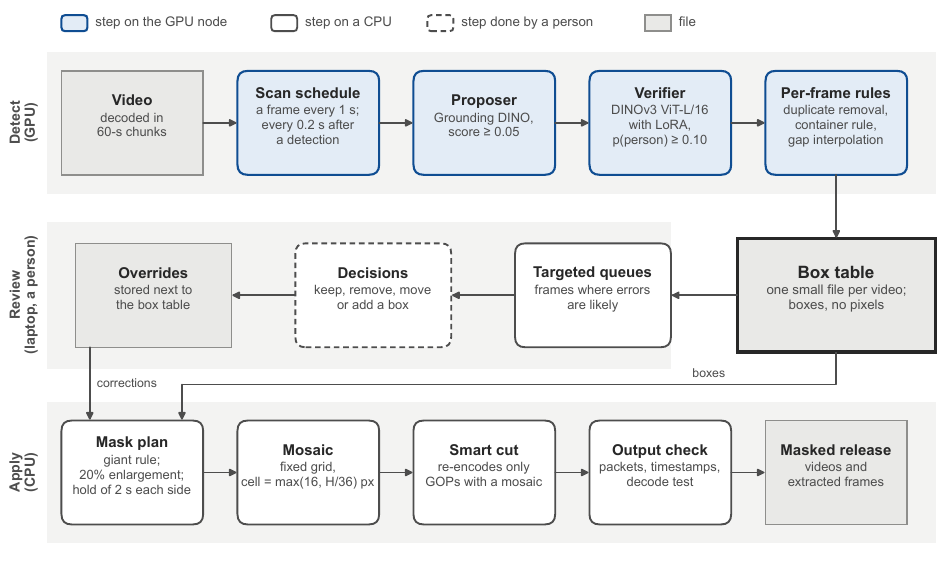}
  \caption{OpenVeil in three lanes around one box table. \emph{Detect} (GPU, once per video): a scan
  schedule selects frames; Grounding DINO proposes boxes with a score of at least 0.05; a DINOv3
  ViT-L/16 crop verifier with a LoRA adapter keeps boxes with \pperson\ \geqs 0.10; per-frame rules and
  gap interpolation tidy the boxes, which are written to one small table per video (boxes, no pixels).
  \emph{Review} (laptop, a person): targeted queues of frames are checked, and the decisions are stored
  next to the table as overrides of the mask plan. \emph{Apply} (CPU): the mask plan (giant rule, 20\%
  enlargement, hold of 2~s on each side) and a fixed-grid mosaic are applied, only the groups of
  pictures that hold a mosaic are re-encoded, and each output is checked against its source. The key
  marks where a step runs and who acts: on the GPU node, on a CPU, or done by a person (dashed
  outline); grey boxes are files.}
  \label{fig:pipeline}
\end{figure}

The work is divided into three commands that can run on different machines (\cref{fig:pipeline}).
\code{openveil detect} runs the models on a GPU and writes one box table per video.
\code{openveil review} lets a person check and correct boxes on a laptop. \code{openveil apply} turns
the tables and the corrections into masked files on central processing units (CPUs) alone. The settings
given in this section are those of the case study (\supptab{1}); \cref{sec:results} gives the
measurements behind them.

\subsection{Detect: propose at a low threshold, verify each crop}
\label{sec:detect}

\begin{figure}[tb]
  \centering
  \includegraphics[width=\linewidth]{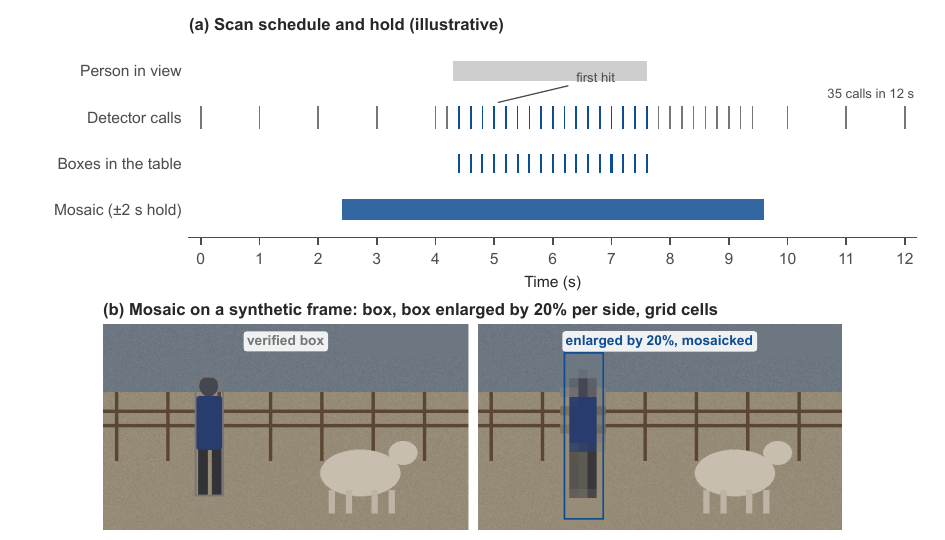}
  \caption{Scan schedule and mosaic. (a)~Scan schedule over 12~s of video with a person in view from
  4.3~s to 7.6~s (illustrative): checkpoints every 1~s; after the first hit the detector scans every
  0.2~s from the last checkpoint without a hit and returns to checkpoints after 2~s without a hit; the
  hold extends the mosaic 2~s on each side of the boxed frames. (b)~The mosaic on a synthetic frame
  (drawn calf shape and fence; no real footage): the verified box (grey, left) and, after masking, the
  box enlarged by 20\% per side with every grid cell that touches it filled with its mean (blue, right),
  made with the OpenVeil mosaic code.}
  \label{fig:schedule}
\end{figure}

\paragraph{Scan schedule.}
People are rare in footage of this kind, so the detector does not look at every frame. By day it scans a
checkpoint frame every 1~s. When a checkpoint holds a person box, the scheduler goes back to the last
checkpoint without one and scans every 0.2~s from there (the dense step), returning to 1-s checkpoints
after 2~s without a hit (\cref{fig:schedule}a). The case study used a \code{clock} policy for the hours
from 23:00 to 06:00, with one checkpoint every 5~s and no dense mode, because people rarely enter the
barns at night; OpenVeil's default policy, \code{always_full}, uses the day schedule at all hours.

\paragraph{Proposer.}
The proposer is Grounding DINO base \citep{liu2023groundingdino}, an open-vocabulary detector that
returns boxes for the phrases of a text prompt. We chose it because, of the detectors compared, only its
low-threshold candidates included nearly every person (\cref{sec:coverage}). The case-study prompt is
\prompt{worker. person. cow. calf.}: \emph{worker} and \emph{person} are the positive phrases, and
\emph{cow} and \emph{calf} give the model somewhere else to put animal-shaped boxes. The score of a box
is its highest positive-phrase score. Every box that scores at least 0.05 becomes a candidate, after
greedy non-maximum suppression (NMS) at an intersection over union (IoU) of 0.6. The threshold is low on
purpose: the proposer should miss almost nothing, and what it adds is left to the verifier. With the
prompt \prompt{person.} alone and a score threshold of 0.45, Grounding DINO found fewer than half of the
labelled people of the validation set (loose recall 0.46; \supptab{15}).
The four-phrase prompt is OpenVeil's default, because the released verifier learned the animal
candidates that it produces as negatives. For footage with other animals or with machines, the
competitor phrases can be replaced (for example \prompt{worker. person. horse.}), after which the
verifier should be checked on reviewed frames (\cref{sec:transfer}).

\paragraph{Verifier.}
The verifier answers a narrower question than a detector: is this crop a person? Each candidate is cut
out with 10\% context on each side, letterboxed to 224\,\by\,224 pixels and scored by a DINOv3 vision
transformer (ViT-L/16) \citep{dosovitskiy2021vit,simeoni2025dinov3} with a rank-32 LoRA adapter
\citep{hu2022lora} and a small multilayer perceptron (MLP) head, which give \pperson. A box is kept when
\pperson\ \geqs 0.10, a low threshold for the same reason as the proposer's. A classifier of this kind
can be trained on the target footage with few positive examples, because a self-supervised backbone
already separates people from animals and a low-rank adapter adjusts it at little cost
\citep{simeoni2025dinov3,hu2022lora}. In the case study the training data were the candidates that the
proposer found in the labelled frames (\cref{sec:sets}): 50,756 crops, of which 1,444 are positives. A
candidate is positive when it covers at least half of a labelled person and that person fills at least
20\% of the candidate box, so that a loose box around a whole group is not taught as correct. The
adapter and the head were trained with the backbone frozen, for 8 epochs with 10 negatives per positive,
which took 35~min on one NVIDIA A10 GPU. The released model is trained on all crops.

\paragraph{Per-frame rules and the box table.}
Two rules tidy the verified boxes of a frame. Of two boxes with IoU~\geqs 0.85, one is removed; and a
box that contains at least two other boxes and is at least three times their union is replaced by that
union (the container rule). Between detector calls, boxes of the same person on both sides of a gap of
up to 3~s by day and 11~s at night are joined by linear interpolation. The result is written to one
compressed table per video, before any margin is added (each step in \suppsec{2}, the training recipe in
\suppsec{3}).

\subsection{Review: a person checks where errors are likely}
\label{sec:review}

\begin{figure}[tb]
  \centering
  \includegraphics[width=\linewidth]{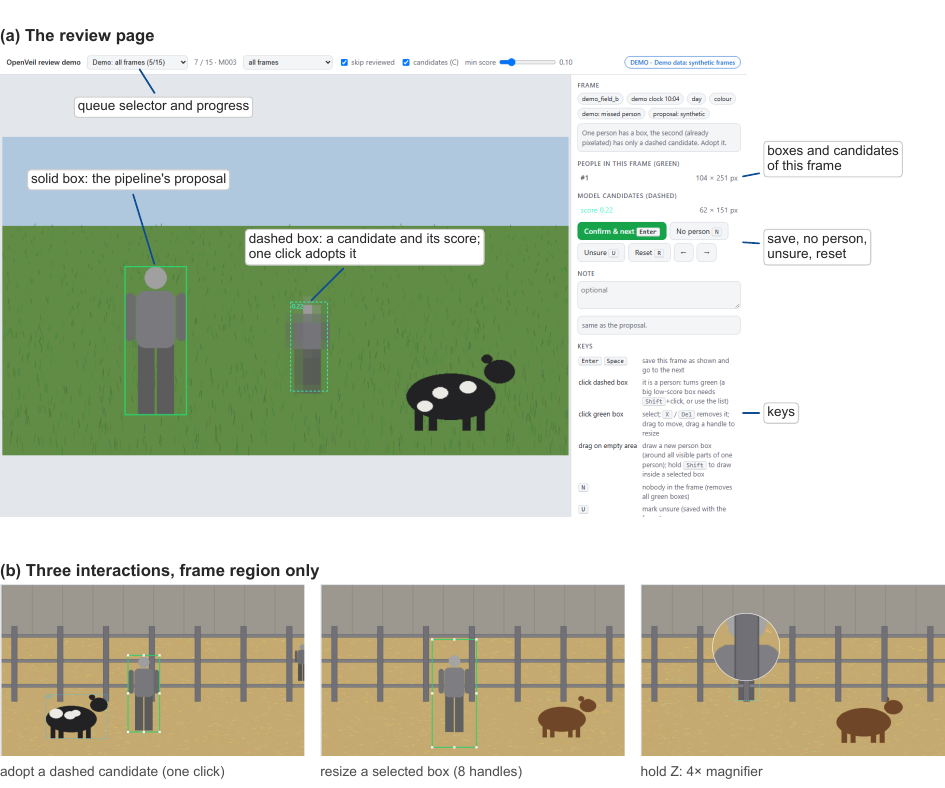}
  \caption{The OpenVeil review tool on synthetic demo data. (a)~The whole page: queue selector, the
  frame with a solid proposal box and a dashed candidate with its score, and the side panel with the
  frame's boxes, candidates, buttons and keys (the keys are listed in \supptab{3}). (b)~Three
  interactions, cropped to the frame: a dashed candidate adopted with one click becomes a box, while a
  low-score candidate on a calf shape stays dashed; a selected box with its eight resize handles; and
  the hold-Z magnifier over a figure half hidden behind a fence. All frames are drawn by
  \texttt{openveil review demo} from synthetic demo data and contain no real person.}
  \label{fig:reviewtool}
\end{figure}

The review tool is a single local web page, served by a small Python server on the reviewer's own
computer; frames never leave that computer. \code{openveil review build} fills queues aimed at the
places where errors are likely: the largest mosaics, where a false alarm costs most; frames in which the
giant rule (\cref{sec:apply}) drops a box; random frames, for an unbiased view; and candidates just
below the verifier threshold. Each frame opens with the pipeline's proposal as solid boxes and the
remaining candidates as dashed boxes (\cref{fig:reviewtool}). The aim is the same for every frame: each
visible person has one box, and nothing else has one. If the proposal is right, Enter saves it and opens
the next frame. A box on an animal is deleted with one key. A missed person gets a box with one click on
a dashed candidate, or by drawing one.

Decisions are appended to a log, and \code{openveil review export} turns them into overrides of the mask
plan. Added boxes are inserted; deleted boxes are removed near their frame; and a moved box is added in
its new place while the old one stays, so that a correction never shrinks the mosaic by accident. The
controls, the export format and the privacy design of the tool are described in \suppsec{4}.

\subsection{Apply: from boxes to masked files}
\label{sec:apply}

\code{openveil apply} computes the mask plan, that is, the boxes to mosaic in each frame, from a box
table. Review corrections are merged first. The giant rule then drops a box that covers at least 55\% of
the image and contains at least two smaller boxes; such a loose group box would otherwise blank most of
the frame. Boxes added by the reviewer are never dropped. Each remaining box is enlarged by 20\% of its
width and height on each side, and every frame receives the boxes of all table rows within 2~s before
and after it (the hold, \pms{2}).

The hold is there because a person is not detected in every frame in which they are visible: in pilot
footage the detections of one person lay up to 3.4~s apart (\suppsec{11}).
Fixed privacy zones over the walkways close such gaps as well, but in the pilot that used three of them
the mosaic covered 32.4\% of all frame area, too much for a dataset that was recorded to show calves, so
OpenVeil has none.
The enlargement rests on the coverage that it adds (\cref{sec:margins}).

The mosaic uses one grid per image, anchored at the top-left corner, with a cell size of
$\max(16, \mathrm{round}(H/36))$ pixels for an image of height $H$, about 36 cells over the image
height; every cell that touches a box is filled with its mean (\cref{fig:schedule}b). Because the grid
is fixed to the image, the mosaic does not flicker when a box moves by a pixel, and it is equally coarse
at every resolution. We did not replace people with generated ones \citep{hukkelas2023deepprivacy2}: the
uses that motivate a release, from behaviour models to video encoders and world models, do not need
synthetic people, and inserted people would add content that was never recorded.

Re-encoding a whole video to change a few per cent of its frames would degrade all the others.
OpenVeil's smart cut re-encodes only the groups of pictures (GOPs) that contain a masked frame, in the
source codec, and copies every other packet unchanged; codecs other than H.264 and HEVC are re-encoded
in full. An output check then confirms that the masked file has the same packet count and timestamps as
the source, that every copied packet is byte-identical and that every re-encoded run decodes to one
frame per packet. A file that fails is re-encoded in full and checked again. Still images extracted from
a video are not sent through the detector: each receives the boxes of its source frame through the same
mask plan (details in Sections~S2.6 and~S2.7).%

\paragraph{Software.}
The review tool needs only the Python standard library, and \code{openveil apply} runs on CPUs with
PyAV. Detection adds PyTorch, Hugging Face Transformers 4.57.1 and PEFT 0.13.2
\citep{mangrulkar2022peft}. The case study used PyAV 18.1.0 \citep{pyav}, whose wheels bundle FFmpeg
8.1.2 \citep{ffmpeg}, and ran on a cloud notebook platform: detection on NVIDIA A10 GPUs with several
worker processes per GPU, and the apply step on CPU nodes.
OpenVeil 1.1.0 reproduces the outputs of the code that masked the case-study dataset (\suppsec{12}) and
adds the changes noted in \supptab{1}.

\section{Case study: masking a calf-behaviour dataset}
\label{sec:casestudy}

\subsection{The footage and the people in it}
\label{sec:footage}

The case study is the public Calf Play Behavior Dataset, recorded in calf barns on two dairy farms in
the Netherlands: a research and teaching farm (Farm~A) and a commercial farm (Farm~B)
\citep{yang2026calfdataset,yang2026calfplay}. It holds 1,173 videos from four recordings (651~h, 50.8
million frames; \cref{tab:recordings}) and 1.1 million JPEG frames extracted at every fifth frame from
the videos of seven sessions.
The cameras hang from the ceiling or stand on poles. They record without audio, in colour by day and in
infrared at night, and burn the date and time into every frame. Farm staff and visitors enter the barns
to feed, clean and check the animals.

\begin{table}[tb]
  \centering
  \caption{The four recordings of the case-study dataset. Duration is the span given by the file names;
  per-recording values are rounded. The mosaic cell size follows the image height (\cref{sec:apply}).
  fps, frames per second.}
  \label{tab:recordings}
  \small
  \setlength{\tabcolsep}{4.5pt}
  \begin{tabular}{@{}llccrrc@{}}
    \toprule
    Recording & Recorded & Resolution (px) & Frame rate (fps) & Videos & Duration (h) & Mosaic cell (px) \\
    \midrule
    \rec{FarmA_R1} & Apr 2024 & 2,304\,\by\,1,296 & 25, variable & 263 & 150.1 & 36 \\
    \rec{FarmA_R2} & Jun 2024 & 1,280\,\by\,720   & 25           & 316 & 169.2 & 20 \\
    \rec{FarmA_R3} & May 2025 & 1,920\,\by\,1,080 & 12           & 288 & 165.9 & 30 \\
    \rec{FarmB_R1} & Jun 2024 & 1,280\,\by\,720   & 25           & 306 & 166.2 & 20 \\
    \midrule
    Total          &          &                   &              & 1,173 & 651.3 &  \\
    \bottomrule
  \end{tabular}
\end{table}

Several things make this footage hard for a person detector. To an open-vocabulary detector, a calf or a
cow looks like a crouching person, and a crouching person like a calf. People are often partial: legs
and boots behind a pen wall, an arm reaching into a pen, a body cut off at the image edge. The cameras
are wide-angle, so a person at the far end of a pen is small. And people are rare, above all at night,
and often in view for only a few seconds.

\subsection{Frame sets and their labels}
\label{sec:sets}

Frames a fraction of a second apart show the same scene, so whole videos, not frames, were assigned to
the training data or to the test data (\cref{tab:sets}). Of the videos of each recording, 10\% (118 in
all) were kept out of the mining of training labels and supplied the test set; clips within 5~min of a
validation clip were kept out as well.

\paragraph{Training set.}
The proposer alone was first run on the key frame at or before every 10-s point of all 1,173 videos
(234,053 frames). From the videos that were not held out we chose 3,004 frames, 751 per recording, from
1,041 videos, in four strata defined by the best proposal of the frame: clearly a person, uncertain
person-like, person-like but scored higher as calf, and random.
Person boxes were pre-labelled by large language model (LLM) agents running Claude Opus~5.5 (Anthropic)
in Claude Code: two agents labelled each frame blind, and a third agent, the arbiter, relabelled the 149
frames on which they disagreed; the two agreed on person versus not person for 98.6\% of the candidate
boxes.
The LLM labels held 502 person boxes in 377 frames. One of the authors then reviewed all 3,004 frames in
the review tool, in the same sessions as the other review queues, and corrected the labels
(\cref{sec:reviewresults}). The tool writes a log line only when a frame is saved, and it logged a
decision for 622 of the frames; the other 2,382, on which the two agents had agreed, needed no change.
One frame was marked unsure and left out of the labels used, which hold 486 people in 361 frames.

\paragraph{Validation set.}
The validation set holds 261 frames from 76 one-minute pilot clips: 145 people in 119 frames, and 142
person-free frames. It was enriched with frames that a stock YOLO detector had missed, so it is harder
than everyday footage. LLM agents pre-labelled it by the same procedure, here with Claude Sonnet~5.5
(Anthropic), and one of the authors then reviewed all 261 frames.
We used it to compare detectors and to set the scan schedule. The 76 source clips, stored at about every
second frame (50,777 frames), are the dense validation clips on which the schedule and the mask plan
were replayed. One of the authors also checked 484 frames of these clips in the review tool: 200 frames from 12
random daytime clips and 284 frames from 12 night clips (\cref{sec:reviewresults}).

\paragraph{Everyday sample and test set.}
From the 118 held-out videos we drew 2,000 frames, the everyday sample. One of the authors checked 731
of them in the review tool and boxed every person; these frames are the test set. They are all 131
frames that a preliminary verifier, trained on the LLM labels before their review, would have masked,
and random draws of 300 of the 721 frames near its threshold and 300 of the 1,148 below it. The test set
holds 172 people in 52 frames, all in colour; none of its 217 infrared frames shows a person.

\paragraph{Cross-fitting.}
The candidate crops of the test set also entered the training of the verifier, so test-set scores had to
come from models that had not seen the frame they scored. The test videos were split into two halves,
and for each half a model was trained on all training-set crops and on the crops of the other half.
Every test frame is thus scored by a model that had not seen its video, although the other half of the
test videos was in that model's training data. All test-set results come from these two cross-fitted
models, which share the recipe of the released verifier but not its weights; we call this the released
configuration. The split is drawn in \suppfig{1}, and \suppsec{5} has the details of each set.

\begin{table}[tb]
  \centering
  \caption{Frame sets, their labels and their use. Training and test data were split by video. The
  validation clips are not held out in this way: the released verifier was trained on other minutes of
  most of the videos from which they come (\cref{sec:margins}). The test set is the human-checked part
  of the everyday sample.}
  \label{tab:sets}
  \footnotesize
  \begin{tabularx}{\linewidth}{@{}>{\raggedright\arraybackslash}p{0.13\linewidth}r>{\raggedright\arraybackslash}X>{\raggedright\arraybackslash}X>{\raggedright\arraybackslash}X>{\raggedright\arraybackslash}X@{}}
    \toprule
    Set & Frames & Source & People & Labels & Used for \\
    \midrule
    Training set & 3,004 & 1,041 videos, 751 frames per recording &
      486 in 361 frames after review (LLM labels: 502 boxes in 377 frames) &
      Claude Opus~5.5 agents; every frame then reviewed by an author &
      Training the verifier; threshold rules \\
    \addlinespace
    Validation set & 261 & 76 one-minute pilot clips & 145 in 119 frames &
      Claude Sonnet~5.5 agents; every frame then reviewed by an author &
      Detector comparison; scan schedule \\
    \addlinespace
    Dense validation clips & 50,777 & the same 76 clips, about every second frame & -- &
      Model output; 484 frames checked by an author & Replay of schedule and mask plan \\
    \addlinespace
    Everyday sample & 2,000 & 118 videos kept out of label mining & -- & -- &
      Weighting of test-set rates \\
    \addlinespace
    Test set & 731 & everyday sample, three strata & 172 in 52 frames &
      One author, in the review tool &
      Evaluation of OpenVeil and of the alternatives; verifier training by cross-fitting \\
    \bottomrule
  \end{tabularx}
\end{table}

\subsection{Coverage and false boxes}
\label{sec:measures}

All methods are scored in the same way. On the test set, a person is \emph{covered} when at least 90\%
of their box lies under the kept boxes after each kept box is enlarged by 20\% per side, as in masking;
IoU recall counts the people with a kept box at IoU~\geqs 0.5. A \emph{false box} is a kept, enlarged
box of which less than 20\% of the area overlaps any person box. We report the share of everyday frames
with at least one false box (false-box frames), the share with a big false box (at least 15\% of the
image) and the share of frame area under false boxes (false area). Rates are weighted back to the 2,000
frames of the everyday sample: a frame from a stratum of $N$ frames of which $n$ were checked has weight
$N/n$. Intervals are 95\% intervals; for weighted rates they come from a bootstrap that resamples frames
within strata (10,000 draws). The test is a single-frame test: it applies the proposer, the verifier and
the per-frame rules to each frame, without the schedule and the hold.

On the validation set the same coverage and false-box rules are applied to the labelled box of each
person; we also report IoU recall and loose recall (any kept box at IoU~\geqs 0.5 with any box drawn on
any part of the person), with Wilson intervals \citep{wilson1927probable}, and the share of the 142
person-free frames that would receive a mosaic. The schedule, the hold, the enlargement and the box
rules were tested by replaying the production code on the dense validation clips.

OpenVeil was compared with five detectors and a face-based baseline: YOLO11x pre-trained on COCO
\citep{jocher2024yolo11,lin2014coco}, a YOLO11l fine-tuned on the training set, OWLv2
\citep{minderer2023owlv2}, LocateAnything-3B \citep{wang2026locateanything}, Grounding DINO without a
verifier, and RetinaFace \citep{deng2020retinaface}. Each had its threshold chosen on the test set in
its favour and is reported at fixed budgets of false-box frames. We also compared seven verifier
variants, trained on the same crops and folds, with each other and with the proposer's score alone, and
set thresholds by rules that do not use the frames on which they are scored (settings in Sections~S6
and~S7).%

\subsection{What the test set can and cannot show}
\label{sec:caveats}

The test set was not built independently of the method that it tests. Its boxes were drawn while the
reviewer checked the proposals of the preliminary verifier, with the Grounding DINO candidates offered
for one-click adoption, and 168 of the 172 boxes are Grounding DINO candidates.
The test set cannot contain a person missed by both the pipeline and the reviewer, and the shapes of its
boxes favour Grounding DINO; for that reason we rank methods by coverage and not by IoU recall. The
verifier is cross-fitted, so half of the test videos were in the training data of each scoring model.
The released threshold of 0.10 and the container rule were chosen with the test set in view. The
training data also favour the verifier over the detector that comes closest to it: YOLO11l was
fine-tuned on the LLM labels before their review and saw no test-set video, whereas the cross-fitted
verifier was trained on the reviewed labels and on half of the test-set videos.

\section{Results}
\label{sec:results}

\subsection{People covered, and the price in false mosaics}
\label{sec:coverage}

Few of the people in this footage are fully visible. Of the 145 people in the validation set, 48 are;
42 show only the lower body, 26 are cut off at the image edge, 26 show only the upper body and 3 only an
arm or a hand.
A face detector has little to work with here. RetinaFace found a face for 66 of the 172 test-set people
at a score of 0.05 and for 33 at 0.5. No face box covered a test-set person, since a face is a small
part of a person's area. A body box drawn from each face covered at most 34 test-set people, with a
false box on 22.6\% of everyday frames (\cref{tab:detectors}), and at most 22 of the 145 validation
people.

\begin{figure}[tb]
  \centering
  \includegraphics[width=\linewidth]{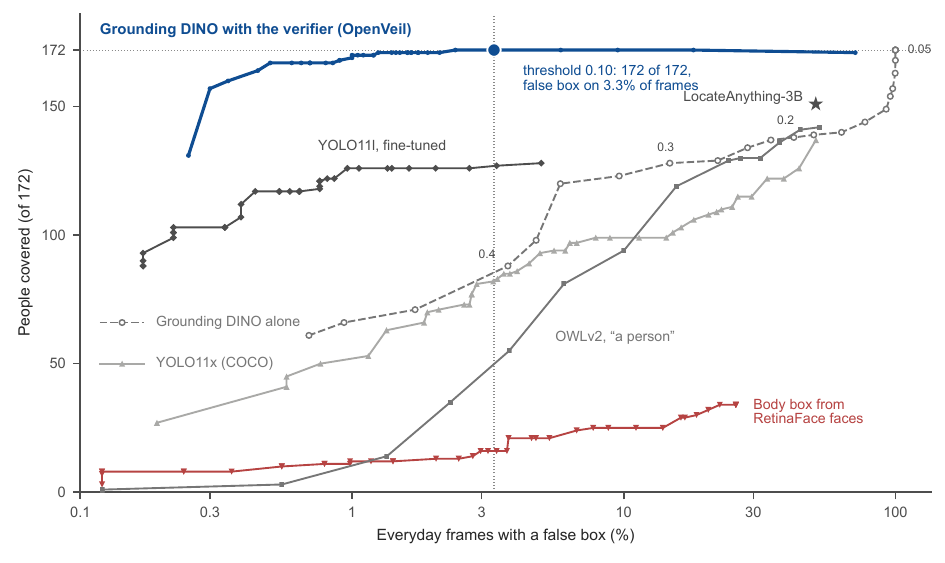}
  \caption{People covered against the price in false boxes on the test set (172 people). Each curve
  sweeps the score threshold of one method; the horizontal axis is the share of everyday frames with a
  false box (log scale; rates weighted to the everyday sample). Only Grounding DINO with the verifier
  (blue; the released threshold of 0.10 is marked) reaches all 172 people at a low false-box rate.
  Grounding DINO alone (dashed; the small numbers are its score thresholds), YOLO11x, the fine-tuned
  YOLO11l and OWLv2 (grey) give up people for fewer false boxes along their curves, LocateAnything-3B
  has a single operating point, and a body box drawn from RetinaFace faces (red) covers at most 34
  people. Thin lines mark 172 people and 3.3\% false-box frames. How the test set was built favours
  methods based on Grounding DINO (\cref{sec:caveats}).}
  \label{fig:coverage}
\end{figure}

In its released configuration, with the cross-fitted verifier (\cref{sec:sets}), OpenVeil covered all
172 people of the test set (\cref{fig:coverage}). It put a false box on 3.3\% [2.6--4.1] of everyday
frames and a big false box on 0.45\% [0.20--0.75], and false boxes took 0.39\% [0.20--0.61] of frame
area. At IoU~0.5, 171 of the 172 people had a matching box. Coverage did not depend on the enlargement:
without it, all 172 people were still covered.
Full coverage of a finite sample still leaves room for misses. With none among 172 people, the one-sided
95\% upper bound on the miss rate is about 1.7\% per person; counted by the 52 frames in which these
people appear, it is about 5.6\% \citep{clopper1934binomial}.

No single detector covered as many people as OpenVeil at the same share of false-box frames. For this
comparison, each detector was run at the threshold that covered the most people while keeping its
false-box frames at or below 3.3\%, the share of OpenVeil. The fine-tuned YOLO11l then covered 126 of
the 172 people, YOLO11x 82, Grounding DINO alone 71 and OWLv2 35 (\cref{tab:detectors}). Grounding
DINO's 71 is its count at a threshold of 0.425, with false
boxes in 1.7\% of frames; the next step of its threshold grid, 0.4, gives 88 people at 3.7\%. At budgets
of 5, 10 and 20\% OpenVeil still covered all 172 people and no single detector more than 128; the
fine-tuned YOLO11l stayed the best alternative, equalled by Grounding DINO alone only at 20\%.
LocateAnything-3B returns no score and so has one operating point, 151 people with a false box on 50.9\%
of everyday frames. The fine-tuned YOLO11l errs on the cautious side. On the validation set it put a box
on only 6.3\% of the person-free frames, but on the test set it covered at most 128 people, even at a
confidence threshold of 0.01. At 0.02, where it covered 127, it boxed at least half of 163 people, so it
mostly misses parts of people, not whole people. The 90\% rule is strict for it, because the test-set
boxes have Grounding DINO's shapes.

\begin{table}[tb]
  \centering
  \begin{threeparttable}
  \caption{Detectors at fixed budgets of false-box frames (test set, 731 frames, 172 people). Covered:
  at least 90\% of a person's box under the kept boxes enlarged by 20\% per side; false-box frames are
  weighted to the everyday sample. The score threshold of each detector was chosen on the test set in
  its favour. How the test set was built favours methods based on Grounding DINO (\cref{sec:caveats}).
  GD, Grounding DINO. Intervals and validation-set results are in Tables~S6 and~S7.}%
  \label{tab:detectors}
  \small
  \begin{tabularx}{\linewidth}{@{}>{\raggedright\arraybackslash}Xl*{4}{>{\centering\arraybackslash}p{1.55cm}}@{}}
    \toprule
     & Most people covered & \multicolumn{4}{c}{People covered at false-box frames of at most} \\
    \cmidrule(l){3-6}
    Method & (false-box frames) & 3.3\% & 5\% & 10\% & 20\% \\
    \midrule
    RetinaFace, face box\tnote{a}       & 0            & 0   & 0   & 0   & 0   \\
    RetinaFace, body box from the face  & 34 (22.6\%)  & 16  & 21  & 25  & 30  \\
    YOLO11x, COCO                       & 137 (51.0\%) & 82  & 93  & 99  & 106 \\
    YOLO11l, fine-tuned on the training set & 128 (5.0\%) & 126 & 128 & 128 & 128 \\
    OWLv2, ``a person''                 & 142 (52.7\%) & 35  & 55  & 81  & 119 \\
    LocateAnything-3B\tnote{b}          & 151 (50.9\%) & --  & --  & --  & --  \\
    GD alone                            & 172 (100\%)  & 71  & 98  & 123 & 128 \\
    \textbf{OpenVeil (GD with the verifier at 0.10)} & \textbf{172 (3.3\%)} & \textbf{172} &
      \textbf{172} & \textbf{172} & \textbf{172} \\
    \bottomrule
  \end{tabularx}
  \begin{tablenotes}\footnotesize
    \item[a] Two people at a score of 0.07 are covered by oversized low-score boxes, not by face boxes;
    face-box coverage is counted as 0 (\supptab{6}).
    \item[b] Returns no score, so there is no threshold to set and no budget below its one operating
    point can be reached.
  \end{tablenotes}
  \end{threeparttable}
\end{table}

A verifier can only remove boxes, so the proposer has to reach near-full coverage at some threshold.
That settled the choice of proposer: it is the detector whose candidates covered the most people at its
lowest threshold. Grounding DINO's candidates at a score of 0.05 covered all 172 test-set people, which
holds partly by construction (\cref{sec:caveats}); all 145 validation people, whose boxes were also
mostly chosen from its candidates; and 481 of the 486 people in the reviewed training labels (99.0\%),
whose frames were selected from the proposer's own scan and whose boxes were mostly adopted from its
candidates as well. We call these 481 the coverable people of the training set.
The next best on the test set, LocateAnything-3B, covered 151 of its people and 105 validation people,
gives no score with which to set a threshold and needed 1.99~s per frame.

Used alone, Grounding DINO can only move along one curve (\cref{fig:coverage}). Keeping every candidate
from 0.05 covers all 172 people and puts a false box on every everyday frame. Raising the threshold to
0.2, 0.3 and 0.4 lowers the false-box frames to 42.3\%, 14.8\% and 3.7\%, and the people covered to 138,
128 and 88.
Its score ranks candidates poorly: over the 9,260 test-set candidates, the area under the receiver
operating characteristic curve (AUC) was 0.713 [0.685--0.743], against 0.994 [0.985--0.998] for the
cross-fitted verifier.
With the verifier, coverage no longer has to be given up to remove false boxes.

\subsection{The verifier and its threshold}
\label{sec:verifier}

\begin{figure}[tb]
  \centering
  \includegraphics[width=\linewidth]{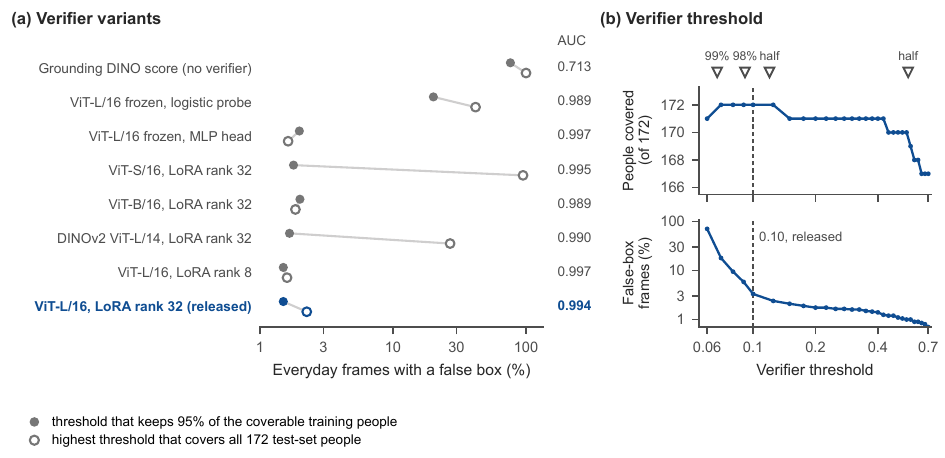}
  \caption{The verifier and its threshold on the test set. (a)~Seven verifier variants and Grounding
  DINO's own score, on the same crops and folds: false-box frames at the threshold that keeps 95\% of
  the coverable people of the training set (filled) and at the highest threshold that still covers all
  172 test-set people, which is set on the test set (open); log scale, with the AUC of each row at the
  right. The released model is in blue. (b)~People covered and false-box frames against the verifier
  threshold in the released configuration; the released threshold of 0.10 is marked, and open triangles
  are thresholds that were not chosen on the frames they are scored on: those that keep 99\% and 98\%
  of the coverable training people, and those chosen on one half of the test videos and applied to the
  other (``half''). Coverage is 172 people from 0.07 to 0.127 and 171 at 0.06 (\suppsec{7}); between
  0.10 and 0.5 it stays at 170--172 while the false-box frames fall from 3.3\% to 1.1\%. Rates are
  weighted to the everyday sample.}
  \label{fig:verifier}
\end{figure}

Seven verifiers, on the same crop table, recipe and two video folds, were compared with each other and
with Grounding DINO's own score (\cref{fig:verifier}a; \supptab{8}). Every DINO-based variant ranked the
candidates far better than that score (AUC 0.989--0.997). At the threshold that kept 95\% of the
coverable people of the training set, Grounding DINO's score covered 142 of the 172 test-set people with
false boxes in 76.3\% of everyday frames; a logistic regression on frozen ViT-L features covered 171
with 20.1\%; and the other variants covered 170--172 with 1.5--2.0\%. The variants separated only when
the threshold had to cover all 172 people. The released ViT-L with a rank-32 adapter then put a false
box on 2.25\% of frames, rank~8 on 1.60\%, ViT-B on 1.85\% and a frozen ViT-L with an MLP head on
1.63\%; DINOv2 ViT-L \citep{oquab2024dinov2} needed 26.9\% and ViT-S 94.9\%, each because of one or two
hard people.

Each variant was trained with one seed, and the comparison does not place the released model ahead of
rank~8, ViT-B or the frozen ViT-L with an MLP head. At the threshold that covers all 172 people, rank~8
and ViT-B left slightly fewer false-box frames than the released model (paired 95\% intervals of the
difference $-1.00$ to $-0.30$ and $-0.75$ to $-0.10$ percentage points), with thresholds set on the
test set; at the 95\% threshold ViT-B left slightly more ($+0.20$ to $+0.85$), and rank~8 and the MLP
head did not differ from it (\suppsec{7}). ViT-B ranked candidates slightly worse than the released
ViT-L (AUC 0.989 against 0.994; paired 95\% interval of the difference $-0.012$ to $-0.0004$) and is
faster, at 574 against 247 crops per second end to end on one A10.
We release ViT-L with a rank-32 adapter because it is the model that masked the case-study dataset. It
was chosen during development, where it left far fewer false boxes than ViT-S with cameras held out
(\supptab{15}), and before this comparison was run. On this test set the adapter was not needed: a
frozen ViT-L with a trained MLP head did as well.

On the test set, coverage was 172 people at thresholds from 0.07 to 0.127 (171 at 0.06, an effect of
the container rule; \suppsec{7}), 171 from 0.15 to 0.425 and 170 at 0.5, while the false-box frames fell
from 3.3\% at 0.10 to 2.4\% at 0.125 and 1.1\% at 0.5 (\cref{fig:verifier}b).
The released 0.10 was set with this curve in view, so we also set thresholds by rules that do not use
the frames on which they are scored (\supptab{9}). On the training set, with out-of-fold scores,
requiring 99\% or 98\% of the coverable people to be covered gave 0.067 and 0.091; both covered all 172
test-set people, with false boxes in 24.9\% and 5.6\% of everyday frames. Requiring every person drove
the threshold to the floor of the score scale, because one or two hard or mislabelled people decide that
rule. The 99\% rule applied to the preliminary verifier gave 0.10 and covered all 172 people, with false
boxes in 8.3\% of frames. A threshold chosen on one half of the test videos and applied to the other
covered 170 of the 172 people [96.9--100\%] with false boxes in 1.6\% [1.1--2.1] of frames, but the
choice is unstable (0.12 in one half and 0.56 in the other), because each half holds only 86 people.
Coverage, then, holds however the threshold is chosen; the false-box rate does not.

\subsection{Margins in space and time}
\label{sec:margins}

\begin{figure}[tb]
  \centering
  \includegraphics[width=\linewidth]{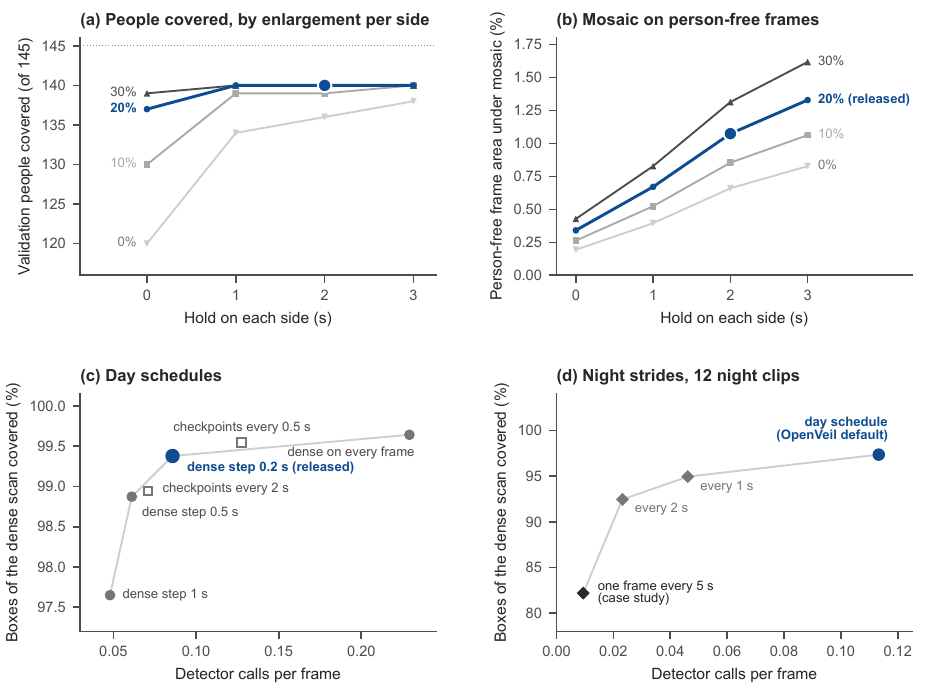}
  \caption{Margins and scan schedule replayed on the dense validation clips. (a)~Validation people
  covered (of 145) against the hold, one line per enlargement. (b)~Mean share of the 142 person-free
  validation frames under mosaic, on the same axes. (c)~Day schedules: share of the boxes of a dense
  scan that the mask plan of a schedule covers, against detector calls per frame. (d)~The 12 night
  clips: sparse strides of 1, 2 and 5~s (the case study used 5~s) and the day schedule, which is
  OpenVeil's default. The boxes of a dense scan are verifier output, not confirmed people. The values
  are in \supptab{10}.}
  \label{fig:margins}
\end{figure}

The rules and margins that follow the verifier were measured on the test set and by replay on the dense
validation clips (\cref{fig:margins}; \supptab{10}). The container rule lowered the frames with a big
false box on the test set from 0.80\% to 0.45\%, at the cost of one IoU match and no coverage. On the
dense validation clips it reduced the mosaic area of the random and night-time frames by 27\% (0.48\%
against 0.66\% of the image), with 140 of the 145 validation people covered either way.

Enlargement and hold complement each other (\cref{fig:margins}a). With neither, the mask plan covered
120 of the 145 validation people; the 20\% enlargement alone covered 137, a \pms{2} hold alone 136, and
both together 140. Beyond 20\% and \pms{1} coverage did not rise further, but the mosaic on person-free
frames kept growing: 0.67\% of the image at \pms{1}, 1.07\% at \pms{2} and 1.33\% at \pms{3}
(\cref{fig:margins}b). We chose \pms{2} because gaps between detections of one person reached 3.4~s in
pilot footage. Where mosaic area matters more than margin, \pms{1} is the better choice: it covered the
same validation people with about 40\% less mosaic on person-free frames. On the 200 human-checked
frames of the random daytime clips, the released setting covered all 27 people; without enlargement it
covered 26, and without enlargement or hold 25.
The giant rule at 0.55 of the image was safe on all 45 reviewed frames at which it fired
(\supptab{11}).

Replayed as a whole on the dense validation clips, with schedule, hold and enlargement, OpenVeil as
released covered 140 of the 145 validation people (0.966 [0.92--0.99]) and found 0.862 [0.80--0.91] of
them under loose matching. These figures are not held out: the released verifier was trained on other
minutes of most of the same videos. The margins have a price on a set that was enriched with hard cases.
In the single-frame test on the validation set, where IoU recall was 0.848 [0.78--0.90], the verifier put
a box on 11.3\% [4.9--19.2] of the 142 person-free frames; after hold and enlargement, 28.2\% of them
received some mosaic, on 1.07\% [0.58--1.67] of their area (\supptab{12}).

The scan schedule kept almost all of what a dense scan finds (\cref{fig:margins}c). Against a
schedule that scans every stored frame in its dense mode, the released schedule used 37\% of the
detector calls (0.086 compared with 0.229 per frame), covered 99.4\% of the boxes found on every stored
frame where that schedule covered 99.6\%, and reached a loose recall of 0.862 against 0.876. Longer
dense steps save more calls but lose boxes: at 1~s the loose recall fell to 0.807, and 4 of the 27
people on the human-checked daytime frames were no longer fully covered.

At night the case study checked one frame every 5~s without a dense mode. Replayed on the 12 night
clips, this covered 82.2\% of the boxes of a dense scan, against 94.9\% at 1~s and 97.4\% for the day
schedule (\cref{fig:margins}d); a 1-s stride would have cost about 20 more Grounding DINO A10-hours for
the 11.5 million night frames (about 25 instead of 5).
The boxes counted here are verifier output, not confirmed people: the review of the 284 frames from
these clips found no person. The 5-s stride was a cost decision of the case study; OpenVeil's default
scans every hour as by day.

\subsection{Human review: boxes kept, removed and added}
\label{sec:reviewresults}

LLM agents pre-labelled the 3,004 training frames and the 261 validation frames, and one of the authors
then reviewed every frame of both sets (\cref{sec:sets}). In the 622 training frames with a logged
decision, the reviewer removed 26 of 359 boxes, added 11 and moved 2. The LLM labels of the training set
were good but not clean: within this review, 17 of 271 boxes (6\%) in a random sample of 200 frames
with people were wrong, most of them on boots next to the pen, and one person had been missed in a
random sample of 300 frames labelled person-free.
The released verifier was trained on the reviewed labels of the training set.

\begin{figure}[tb]
  \centering
  \includegraphics[width=\linewidth]{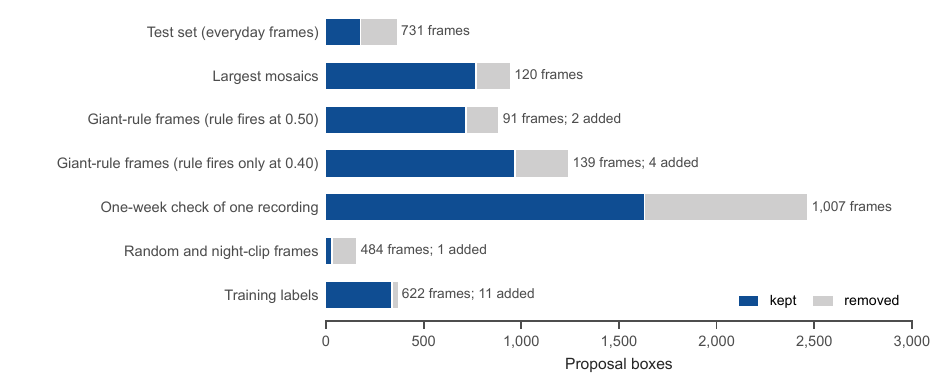}
  \caption{Proposal boxes kept (blue) and removed (grey) in each review queue; the number of frames and
  of boxes added by hand is given at the end of each bar. The reviewer mostly removed boxes and added
  18 in 3,194 frames. Removed boxes are false alarms on calves, objects and loose group boxes; in the
  giant-rule queues they include the dropped giant boxes, and in the training-label queue they are
  wrong LLM labels. Counts are the logged decisions.}
  \label{fig:reviewoutcome}
\end{figure}

Over all queues, the review tool logged decisions for 3,194 frames in about 2.4~h of active work; the
median was 0.6~s per frame, and a few frames took much longer (\cref{fig:reviewoutcome}; \supptab{13}).
The 6,362 proposal boxes became 4,617 boxes after review: 4,599 were kept, 1,763 removed as false alarms
on calves, objects and loose group boxes, 38 moved and 18 added. Of the 18 boxes added over all queues,
11 corrected LLM training labels, 6 were drawn in giant-rule frames and 1 was a further candidate box
adopted in a daytime frame.
No person was missing in the test set: all 172 people were already covered by the proposal, and the
reviewer adjusted 2 boxes for tighter coverage. The review found no missing box in the largest-mosaic
queue either, nor in the one-week check of one recording, in which 155 of the 734 intervals with a
mosaic held a person and the others were brief false mosaics.

\subsection{The released dataset and what it cost}
\label{sec:release}

\begin{figure}[tb]
  \centering
  \includegraphics[width=\linewidth]{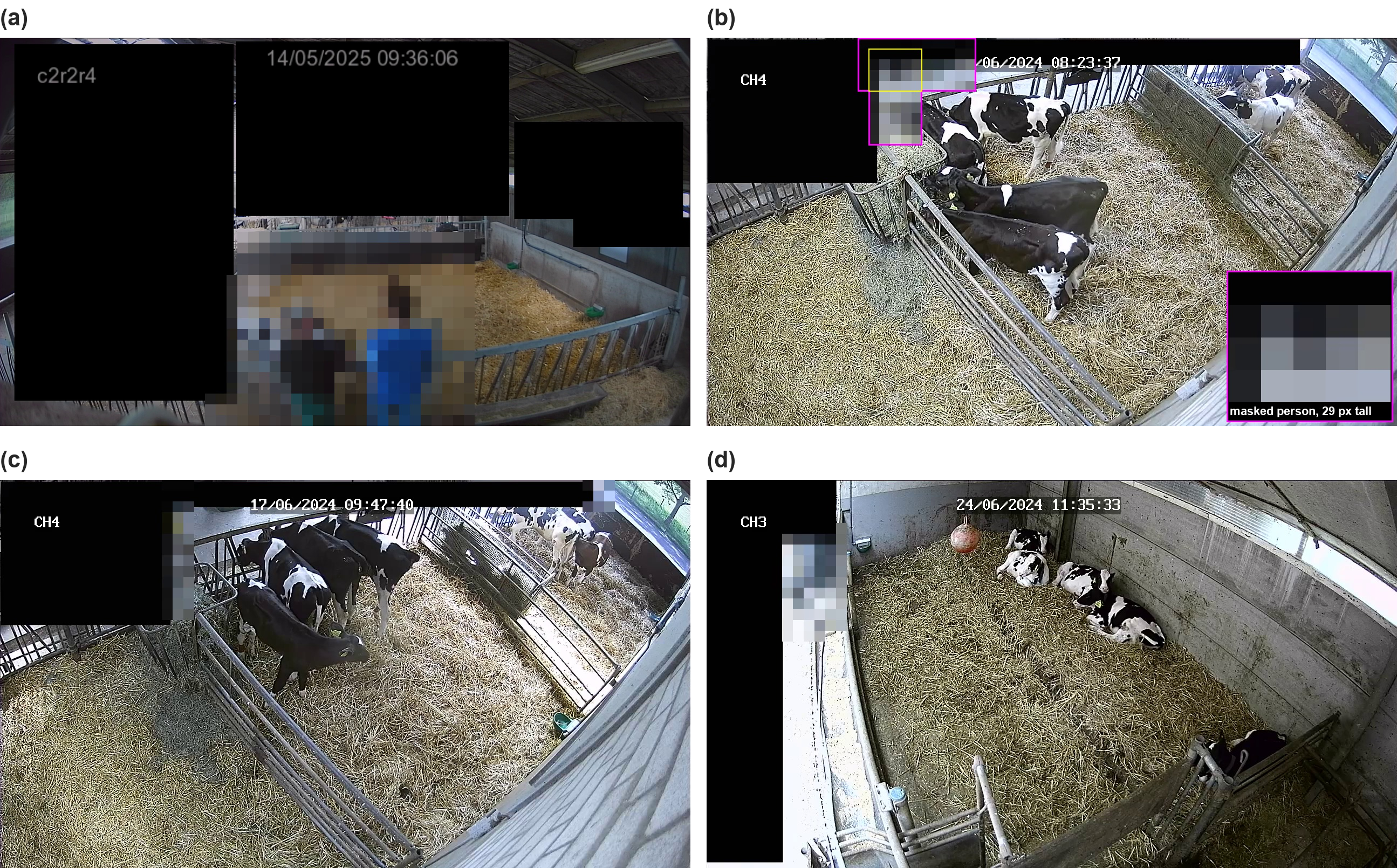}
  \caption{Masked frames of the public release. (a)~FarmA\_R3: two people masked (1,920\,\by\,1,080
  frame, 30-px cells). (b)~FarmA\_R2: a partly visible person (lower body) of 40\,\by\,29~px (0.13\% of
  the frame), found and masked; the inset enlarges the outlined mosaic three times. (c)~FarmA\_R2: a
  partly visible person of 29\,\by\,58~px (0.18\% of the frame). (d)~FarmB\_R1: a partly visible person
  of 47\,\by\,84~px (0.43\% of the frame). All panels are full frames of the public masked release,
  approved in writing by the data owners; in each, a reviewer had confirmed that every person is covered
  and added no box.}
  \label{fig:examples}
\end{figure}

The reviewed mask plan was applied to all videos and to the tar archives of JPEG frames on CPU nodes.
Of the 1,173 videos, 1,023 received a mosaic, in 3,439,052 of their 44,717,570 frames (7.7\%). That is
6.8\% of all video frames, from 4.6\% in \rec{FarmA_R1} to 11.1\% in \rec{FarmA_R3} (\supptab{14}).
Before hold and enlargement, 4.67\% of all frames received a box from the pipeline, false alarms
included.
The other 150 videos had no detection and are byte-identical to the source. Smart cut re-encoded 12.0\%
of the packets of the masked files and copied the other 88.0\% unchanged; the 33 masked files in MPEG-4
Part~2 were re-encoded in full. Every masked file passed the output check, and no fallback re-encode was
needed. Of the 1,106,110 JPEG frames, 129,011 (11.7\%) carry a mosaic. The masked dataset was released
as version 1.2 on 13 November 2026 \citep{yang2026calfdataset}; \cref{fig:examples} shows four of its
frames.

\begin{table}[tb]
  \centering
  \caption{The release in numbers: compute, files and review. The detector was called on 5\% of the
  frames, and 88\% of the packets of the masked videos were copied unchanged. GPU hours are estimates
  for one NVIDIA A10 at 5.9 frames per second (Grounding DINO) and 160 crops per second (verifier),
  scaled to the 50.8 million video frames; verifier hours assume as many candidates per call as in the
  dense validation clips (12.0 per frame).}
  \label{tab:release}
  \small
  \begin{tabularx}{\linewidth}{@{}>{\raggedright\arraybackslash}p{0.24\linewidth}>{\raggedright\arraybackslash}X>{\raggedright\arraybackslash}X@{}}
    \toprule
    Item & Case study & Alternatives \\
    \midrule
    Detector calls per frame & 0.0508 (day schedule by day, 5-s stride at night) &
      0.086 (day schedule at all hours); 0.229 (dense on every stored frame); 1 (every frame) \\
    Grounding DINO, A10-hours & about 122 & about 205; 548; 2,391 \\
    Verifier, A10-hours & about 54 & about 91; 243; -- \\
    Verifier training & 35~min on one A10 & -- \\
    \addlinespace
    Videos with a mosaic & 1,023 of 1,173; the other 150 are byte-identical to the source & -- \\
    Frames carrying a mosaic & 6.8\% of video frames; 11.7\% of JPEG frames & -- \\
    Video packets re-encoded & 12.0\% of the packets of the 1,023 masked videos &
      100\% with a full re-encode \\
    \addlinespace
    Human review & 3,194 frames with a logged decision in about 2.4~h (median 0.6~s per frame) &
      more than 8,000~h to look once at every frame at 0.6~s \\
    \bottomrule
  \end{tabularx}
\end{table}

Because people are rare, the detector ran on few frames (\cref{tab:release}). It was called 2,582,574
times, 0.0508 times per frame: 2.47 million calls for 39.3 million day frames and 0.11 million for 11.5
million night frames. GPU time was not metered per step, so we estimate it from throughput: 5.9 frames
per second for Grounding DINO and 160 crops per second for the verifier on one A10, a conservative
figure for a verifier that ran at 247 crops per second end to end (\supptab{8}). The call count then
corresponds to about 122 Grounding DINO A10-hours. Running the day schedule at all hours, OpenVeil's
default, would need about 296 A10-hours (205 for the proposer and 91 for the verifier), and running the
proposer alone on every frame 2,391.

\section{Discussion}
\label{sec:discussion}

\subsection{Finding people and deciding who is a person}
\label{sec:shows}

The case study shows why releasing working footage is not a matter of finding a better threshold. A
detector's single score cannot both find every person and tell a crouching worker from a calf. Grounding
DINO had almost every person among its candidates, but at a threshold that spared the calves it lost
most of the people. A classifier that sees one crop at a time and is trained on the target footage can
tell the two apart. With it, OpenVeil covered all 172 people of the test set; Grounding DINO alone,
limited to the same share of false-box frames, covered 71. Faces could not have done this work: a face
detector found a face for 66 of the 172 people, so
only whole-body boxes could hide them. Which backbone does the verifying matters less than having a
verifier at all: every DINO-based variant removed most false boxes, and ViT-B, a rank-8 adapter and a
frozen ViT-L with an MLP head did about as well as the model we released. Proposing and then classifying
is an old pattern \citep{viola2001cascade,girshick2014rcnn}. What has changed is that a self-supervised
backbone with a small head trained on 1,444 positive crops, with or without an adapter, is enough to
make it work on footage that no public detector was trained on.

Storing boxes moved several decisions to after detection. The giant-rule threshold, the hold, the
enlargement and every correction by the reviewer were set once the box tables existed and were applied
without a GPU, so each could be checked, changed and applied again at low cost. The reviewer, in turn,
verified boxes and rarely had to draw one. With queues aimed at large mosaics, rule firings, random
frames and uncertain candidates, the work was mostly the removal of boxes, which is how 3,194 logged
frames fitted into 2.4~h.

\subsection{Taking OpenVeil to other footage}
\label{sec:transfer}

What carries over to other footage is the code and the procedure. The released weights are only a first
guess there: the verifier has seen calf-barn crops only, and the default prompt names cows and calves as
the competitors. For a new dataset we suggest the following procedure.
\begin{enumerate}
  \item Run detection with the released verifier and the \code{always_full} policy, which scans every
  hour as by day.
  \item Review the queues that OpenVeil builds: largest mosaics, random frames and below-threshold
  candidates.
  \item If the review finds false alarms or misses, set a new verifier threshold on the reviewed frames
  or train a new adapter on them, which took 35~min in the case study.
  \item Apply the mask plan and the review corrections, and keep a random sample checked by a person as
  evidence for the dataset card.
\end{enumerate}
The threshold has to be set on the dataset's own reviewed frames, because the false-box rate it yields
does not carry over (\cref{sec:verifier}), and frames used to evaluate the verifier must be kept out of
its training.

Most of the compute goes to the proposer, so the scan schedule and the day/night policy are the first
things to adjust. Where verifier time matters, a ViT-B adapter is a reasonable choice: it ran 2.3 times
faster and, with its own threshold, left a similar share of false-box frames (\cref{sec:verifier}).
Smart cut keeps most of a release byte-identical to the source, which matters to users who compare
masked footage with annotations made on the original.

\subsection{Limits of the evidence}
\label{sec:limits}

OpenVeil has been tested on four recordings from two farms with similar barns and cameras. Its coverage
figure counts the people whom the pipeline or the reviewer found (\cref{sec:caveats}): a person missed
by both is in no count, and reviewing detections cannot reveal people who were never detected. The 172
people of the test set appear in only 52 frames, all in colour, which is why the figure comes with an
upper bound (\cref{sec:coverage}). None of the 217 infrared frames of the test set and none of the 284
reviewed frames of the night clips shows a person, so coverage in infrared could not be measured. In the
case-study run, gap interpolation did not cross the 60-s chunk boundaries at night, which left 3,216
frames at 134 boundaries outside the hold (\suppsec{10}); the boxes on either side of these gaps are
verifier output, and OpenVeil now closes the gaps.

The false-box rate depends on how the threshold is set. Thresholds that were not chosen on the frames
they were scored on covered 170 to 172 of the 172 people, but with false boxes in anything from 1.5\% to
100\% of frames (\supptab{9}). The released model, trained on all data, has no held-out estimate.
The evaluation on the test set is a single-frame test. The schedule, the hold and the enlargement were
tested in two ways. The first was the replay on the dense validation clips, where the reference was
mostly the author-reviewed validation labels or model output (the boxes of a dense scan). The second was
the review of the largest-mosaic queue and the one-week check (\cref{sec:reviewresults}).
The verifier variants were trained with one seed each, and YOLO11l, the one detector trained on this
footage, saw fewer, unreviewed labels.

OpenVeil masks pixels only: it does not handle audio, metadata or burned-in text. And no method can
guarantee that no person remains in millions of frames.

The most useful next test is a second domain, such as factory or laboratory video, where machines and
not animals are what a detector takes for people; it would show how many reviewed frames a new verifier
needs. The below-threshold queue of the review tool already collects the candidates about which the
verifier is least sure; once reviewed, they are the natural next training examples. Inside verified
boxes, a promptable segmenter \citep{ravi2024sam2} could replace the box by the outline of the person,
so that less of the animals and the scene is hidden.

\section{Conclusion}
\label{sec:conclusion}

We argue that hiding people in footage recorded to show animals or objects calls for a division of
labour: a detector proposes person boxes broadly, a verifier narrows the proposals afterwards, and a
person reviews where errors are likely. Our reason is that a detector has a single score threshold. A
high threshold misses partly hidden people, and a low one puts false boxes on those animals or objects.
With two models, the proposer only has to find almost every person, and the verifier only has to decide
whether one crop shows a person, which it can learn from reviewed frames of the footage at hand.

The evidence is one case study, the masking of a calf play-behaviour dataset of 1,173 videos. On the
test set, OpenVeil's proposer and verifier covered all 172 people at the price of a false box on 3.3\%
of everyday frames. A single detector that was limited to the same share of false-box frames covered
fewer people: the best one covered 126, and the proposer without the verifier covered 71. A person
missed by both the pipeline and the reviewer is not counted in these figures. The logged review of
3,194 frames took about 2.4~h. For other footage, the released weights are only a starting point: a new
dataset needs its own reviewed frames and its own threshold.

\FloatBarrier
\section*{Supplementary Material}

The Supplementary Material accompanies this article. It gives the settings of the case study
(\suppsec{1}), the pipeline in detail (\suppsec{2}), the training labels and the training of the
verifier (\suppsec{3}), the review tool (\suppsec{4}), the frame sets and measures (\suppsec{5}), the
detector comparison (\suppsec{6}), the verifier variants and threshold rules (\suppsec{7}), the margins,
box rules and scan schedule (\suppsec{8}), the human review per queue (\suppsec{9}), the release per
recording (\suppsec{10}), the earlier approaches (\suppsec{11}), the agreement of the released package
with the case-study code (\suppsec{12}) and the licences (\suppsec{13}).

\section*{Author contributions}

\textbf{Haiyu Yang:} conceptualisation, methodology, software, formal analysis, investigation, data
curation, validation, visualisation, writing (original draft; review and editing).
\textbf{Miel Hostens:} supervision.

\section*{Competing interests}

The authors declare no competing interests.

\section*{Data availability}

The masked case-study dataset is public at
\url{https://huggingface.co/datasets/Sonam5/Calf-Play-Behavior-Dataset} (version 1.2, CC0-1.0,
\href{https://doi.org/10.57967/hf/11689}{doi:10.57967/hf/11689}) \citep{yang2026calfdataset}. The
recordings and their annotations are described in a data descriptor \citep{yang2026calfplay}. The
unmasked footage, the training crops and frames, and the evaluation frames are not released.

\section*{Code availability}

OpenVeil is available at \url{https://github.com/Sonam525/OpenVeil} under the Apache-2.0 licence, with
the review tool and its synthetic demo, the presets, documentation and a test suite on synthetic data.
This article describes OpenVeil 1.1.0, whose default prompt is the case-study prompt
(\cref{sec:detect}). That version is archived on Zenodo,
\href{https://doi.org/10.5281/zenodo.24506134}{doi:10.5281/zenodo.24506134} \citep{yang2026openveil};
the concept DOI \href{https://doi.org/10.5281/zenodo.24505557}{10.5281/zenodo.24505557} covers all
versions. The verifier weights (LoRA adapter and head) are available at
\url{https://huggingface.co/Sonam5/OpenVeil-Verifier-DINOv3-ViT-L16} under the DINOv3 License
\citep{yang2026verifier}; the DINOv3 backbone has to be obtained from Meta under the same licence. The
licences of the dependencies and of the baselines are listed in \suppsec{13}.

\section*{Ethics statement}

The farmers consented to the cameras and to continuous recording and were told that they might appear on
video; the farm owners consented to the public release of the data. The unmasked footage, the crops and
the evaluation frames are not released. All review of frames was done by one of the authors on a local
machine. The example frames in this article are masked frames of the public release, approved in writing
by the data owners, and no unmasked image of a person appears in the article or its Supplementary
Material. The recordings come from an observational study of the animals that was approved by the Animal
Welfare Body of Utrecht University (approval 10818-2024-02) and is described in the data descriptor
\citep{yang2026calfplay}.

\section*{Use of AI tools}

AI tools were used in two ways. For labelling, agents running Claude Opus~5.5 (Anthropic) in Claude Code
pre-labelled the 3,004 training frames, and agents running Claude Sonnet~5.5 pre-labelled the 261
validation frames, in October 2026 (\cref{sec:sets}); one of the authors then reviewed every frame of
both sets. For code, Claude in Claude Code was used to write and test the software and the figure
scripts. The authors checked the labels and the code and are responsible for the content of the article.

\bibliographystyle{plainnat}
\bibliography{refs}

\end{document}

%% file: preamble.tex
\usepackage[T1]{fontenc}
\usepackage[utf8]{inputenc}
\usepackage{textcomp}%
\usepackage{lmodern}
\usepackage[a4paper,margin=2.5cm]{geometry}
\usepackage{microtype}
\usepackage{amsmath,amssymb}
\usepackage{graphicx}
\graphicspath{{figures/}}
\usepackage{xcolor}
\usepackage{array}
\usepackage{booktabs}
\usepackage{tabularx}
\usepackage{longtable}
\usepackage{makecell}
\usepackage{threeparttable}
\usepackage[font=small,labelfont=bf,labelsep=period]{caption}
\usepackage{placeins}
\usepackage{enumitem}
\setlist{nosep,leftmargin=*}
\usepackage{authblk}

\usepackage[round,authoryear,sort]{natbib}
\defcitealias{gdpr2016}{GDPR, 2016}%
\usepackage{lineno}%
\usepackage{xurl}
\usepackage[colorlinks=true,linkcolor=blue!50!black,citecolor=blue!50!black,urlcolor=blue!50!black]{hyperref}
\usepackage{orcidlink}
\hypersetup{pdfauthor={Haiyu Yang, Miel Hostens}}%
\usepackage[capitalise,nameinlink]{cleveref}
\crefname{section}{Section}{Sections}
\Crefname{section}{Section}{Sections}

\newcommand{\doi}[1]{\href{https://doi.org/#1}{doi:\nolinkurl{#1}}}

\DeclareRobustCommand{\code}[1]{\texttt{\frenchspacing\detokenize{#1}}}%
\DeclareRobustCommand{\prompt}[1]{\mbox{\code{#1}}}%
\newcommand{\pperson}{$p(\mathrm{person})$}
\newcommand{\pms}[1]{$\pm$#1~s}%
\newcommand{\by}{$\times$}%
\newcommand{\geqs}{$\geq$\,}

\newcommand{\rec}[1]{\mbox{\detokenize{#1}}}%
\newcommand{\suppsec}[1]{Section~S#1}
\newcommand{\supptab}[1]{Table~S#1}
\newcommand{\suppfig}[1]{Fig.~S#1}

\newcommand{\papertitle}{OpenVeil: detect, verify and mosaic people in video datasets before release, with an interactive human review tool}
\newcommand{\paperauthors}{%
  \author[1,*]{Haiyu Yang\,\orcidlink{0009-0007-5125-4958}}
  \author[1]{Miel Hostens\,\orcidlink{0000-0001-5376-976X}}
  \affil[1]{Department of Animal Science, College of Agriculture and Life Sciences, Cornell University, Ithaca, NY, USA}
  \affil[*]{Corresponding author: \href{mailto:hy625@cornell.edu}{hy625@cornell.edu}}}